\documentclass{midl} 

\usepackage{mwe} 
\jmlrvolume{Under Review}
\jmlryear{2020}
\jmlrproceedings{~}{~}

\title[Smart Chest X-ray Worklist Prioritization using AI]{Smart Chest X-ray Worklist Prioritization using Artificial Intelligence: A Clinical Workflow Simulation}

\midlauthor{\Name{Ivo M. Baltruschat\nametag{$^{1,2,3}$}} \Email{i.baltruschat@uke.de}\\
	\addr $^{1}$ Department for Diagnostic and Interventional Radiology and Nuclear Medicine,
	University Medical Center Hamburg-Eppendorf, Hamburg, Germany \\
	\addr $^{2}$ Institute for Biomedical Imaging, Hamburg University of Technology, Hamburg, Germany \\
	\addr $^{3}$ DAISYlabs, Forschungszentrum Medizintechnik Hamburg, Hamburg, Germany \AND
	\Name{Leonhard Steinmeister\nametag{$^{1}$}}\\
	\Name{Hannes Nickisch\nametag{$^{4}$}}\\
	\addr $^{3}$ Philips Research, Hamburg, Germany \\
	\Name{Axel Saalbach\nametag{$^{4}$}}\\
	\Name{Michael Grass\nametag{$^{4}$}}\\
	\Name{Gerhard Adam\nametag{$^{1}$}}\\
	\Name{Tobias Knopp\nametag{$^{1,2}$}}\\
	\Name{Harald Ittrich\nametag{$^{1}$}}\\
}

\begin{document}

\maketitle

\begin{abstract}
\paragraph*{Objective}
The aim is to evaluate whether smart worklist prioritization by artificial intelligence (AI) can optimize the radiology workflow and reduce report turnaround times (RTAT) for critical findings in chest radiographs (CXRs). Furthermore, we investigate a method to counteract the effect of false negative predictions by AI –- resulting in an extremely and dangerously long RTAT, as CXRs are sorted to the end of the worklist.

\paragraph*{Methods}
We developed a simulation framework that models the current workflow at a university hospital by incorporating hospital specific CXR generation rates, reporting rates and pathology distribution. Using this, we simulated the standard worklist processing ``first-in, first-out'' (FIFO) and compared it with a worklist prioritization based on urgency. Examination prioritization was performed by the AI, classifying eight different pathological findings ranked in descending order of urgency: pneumothorax, pleural effusion, infiltrate, congestion, atelectasis, cardiomegaly, mass and foreign object. Furthermore, we introduced an upper limit for the maximum waiting time, after which the highest urgency is assigned to the examination.

\paragraph*{Results}
The average RTAT for all critical findings was significantly reduced in all Prioritization-simulations compared to the FIFO-simulation (e.g. pneumothorax: 35.6 min vs. 80.1 min; p $<0.0001$), while the maximum RTAT for most findings increased at the same time (e.g. pneumothorax: 1293 min vs 890 min; p $<0.0001$). Our ``upper limit'' substantially reduced the maximum RTAT all classes (e.g. pneumothorax: 979 min vs. 1293 min / 1178 min; p $<0.0001$).

\paragraph*{Conclusion}
Our simulations demonstrate that smart worklist prioritization by AI can reduce the average RTAT for critical findings in CXRs while maintaining a small maximum RTAT as FIFO.
\end{abstract}


\section{Introduction}
Growing radiologic workload, shortage of medical experts and declining revenues often lead to potentially dangerous backlogs of unreported examinations, especially in publicly funded health care systems. With the increasing demand for radiological imaging, the continuous acceleration of image acquisition and the expansion of teleradiological care, radiologists are nowadays working under increasing time pressure , which cannot be compensated by improving RIS-PACS integration or use of speech recognition software [1].

Delayed communication of critical findings to the referring physician bears the risk of delayed clinical intervention and impairs the outcome of medical treatment [2-4], especially in cases requiring immediate action, e.g. tension pneumothorax or misplaced catheters. For this reason, the Joint Commission defined the timely reporting of critical diagnostic results as an important goal for patient safety [5]. 
Many institutions still process their examination worklists following the first-in, first-out (FIFO) principle. The ordering physician’s urgency information is often incomplete or presented as ambiguous and ill-defined priority level, such as ``critical'', ``ASAP'' or ``STAT'' [6-7].

Artificial intelligence (AI) methods such as convolutional neural networks (CNNs) offer promising options to streamline the clinical workflow. Automated disease classification can enable real-time prioritization of worklists and reduce the report turnaround time (RTAT) for critical findings [8]. For chest X-rays (CXRs) a potential benefit of real-time triaging by CNNs has been reported by Annarumma et al. [9], but they focused mostly on the development of an AI system without a real clinical simulation and does not present the maximum RTAT.

The benefits of smart worklist prioritization need to be discussed not only on the basis of the average RTAT but also of the maximum RTAT. One problem with using AI methods for smart worklist prioritization is that it can and does happen that a critical finding is ``overlooked'' by the AI -- i.e. the false negative rate (FNR) is not zero. The higher the FNR, the more likely it is that individual examinations with critical findings will be sorted to the end of the worklist, risking delayed treatment.

In this work, we simulate multiple smart worklist prioritization methods for CXR in a realistic clinical setting, using empirical data from a university hospital. We develop a realistic simulation framework and evaluate whether AI can reduce RTAT for critical findings by using smart worklist prioritization instead of the standard FIFO sorting. Furthermore, we propose a thresholding method for the maximum waiting time to reduce the effect of false negative predictions by AI.

\section{Material \& Methods}
\subsection{Convolutional Neural Network Architecture and Training}
Based on previous work [10,11] we employed a tailored ResNet-50 architecture with a larger input size of 448 x 448. Furthermore, we preprocessed each CXR using two methods before training (i.e. lung field cropping and bone suppression), as shown in earlier experiments the highest average AUC value is achieved by combining both methods in an ensemble [11]. We pre-trained our model on the publicly available ChestX-ray14 dataset [12] and, after replacing the last dense layer of the converged model, we fine-tuned it on the opensource OpenI dataset [13]. Two expert radiologists from our department annotated a revised OpenI dataset (containing 3125 CXRs) regarding eight findings [11]: pneumothorax, congestion, pleural effusion, infiltrate, atelectasis, cardiomegaly, mass, foreign object. 
Due to the importance of pneumothorax detection and the low number of cases with ``pneumothorax''(n=11) in OpenI, we used the specifically trained and adapted ResNet-50 model of Gooßen et al. [14] for this finding. Our final model therefore includes two separate CNNs, both of which obtained the highest average AUC value (Figure \ref{fig:ROC}) in previous experiments compared to different network architectures. The processing time per image with a Nvidia GTX 1080 GPU is around 10ms and adds a negligible overhead to the reporting process.

\begin{figure}[htbp]
	\floatconts
	{fig:ROC}
	{\caption{Receiver operating characteristics of the artificial intelligence algorithm for all eight different findings.}}
	{\includegraphics[width=1.0\linewidth]{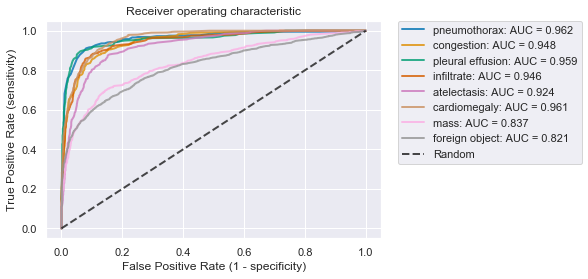}}
\end{figure}

\subsection{Pathology Triage}
For triage, a ranking of the pathologies was defined by two experienced radiologists (Table \ref{tab:Finding}), reflecting the urgency for clinical action. As our annotations did not include different degrees of pathology manifestation, only the presence of a pathology was considered for the prioritization. Furthermore, the impact of different pathology combinations was not considered.

\subsection{Workflow Simulation}
\begin{figure}[htbp]
	\floatconts
	{fig:Overview}
	{\caption{Workflow Simulation. A chest X-ray (CXR) machine is constantly generating CXRs. To each CXR, zero or up to eight findings are assigned. CXRs are either sorted into the worklist chronologically (FIFO) or according to the urgency based on the prediction by artificial intelligence (PRIO). Finally, worklists are processed by a virtual radiologist.}}
	{\includegraphics[width=1.0\linewidth]{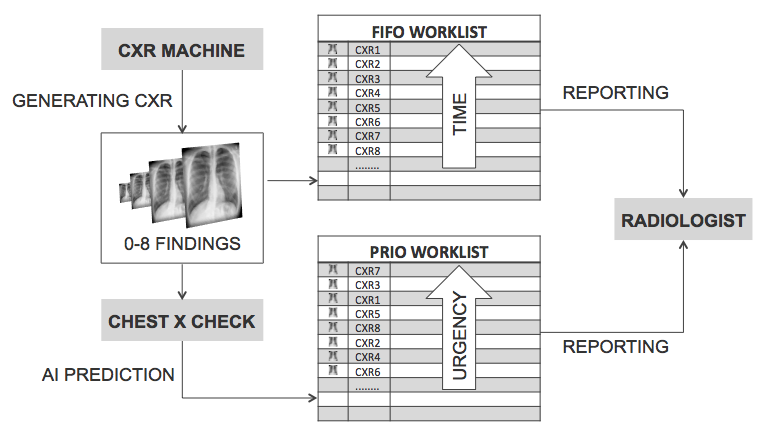}}
\end{figure}

To evaluate the clinical effect of a CXR worklist rearrangement by AI under realistic conditions, we analyzed the current workflow in a radiology department at a university hospital and transferred this data into a computer simulation (Figure \ref{fig:Overview}). We designed a model consisting of four main parts. First, a discrete distribution of how often CXRs are generated. Secondly, the department specific disease prevalence for eight findings to assign labels to the CXRs. Thirdly, the performance for all eight findings of a state-of-the-art CNN to classify each CXR. Fourthly, a second discrete distribution of how fast a radiologist finalizes a CXR report.

By monitoring the CXR acquisition and reporting process of 1408 examinations in total, we were able to extract a discrete distribution of the acquisition and reporting times of subsequent CXRs to calculate the RTAT [15]. The department specific distribution of pathologies was analyzed by manually annotating all eight findings in 600 CXRs.

To simulate the clinical workload throughout the day, a model of a CXR machine was developed, constantly generating new examinations which fill up a worklist. The generation process was modeled using the discrete distribution of our acquisition time analysis (Figure \ref{fig:DistributionGeneration}) including all effects, such as different patient frequency during day and night.
\begin{figure}[htbp]
	\floatconts
	{fig:DistributionGeneration}
	{\caption{Discrete distribution of chest X-ray generation speed. The X-axis shows the day time in 24-hour format and the Y-axis shows the calculated time deltas. The histogram in X- and Y- direction is shown in green.}}
	{\includegraphics[width=1\linewidth]{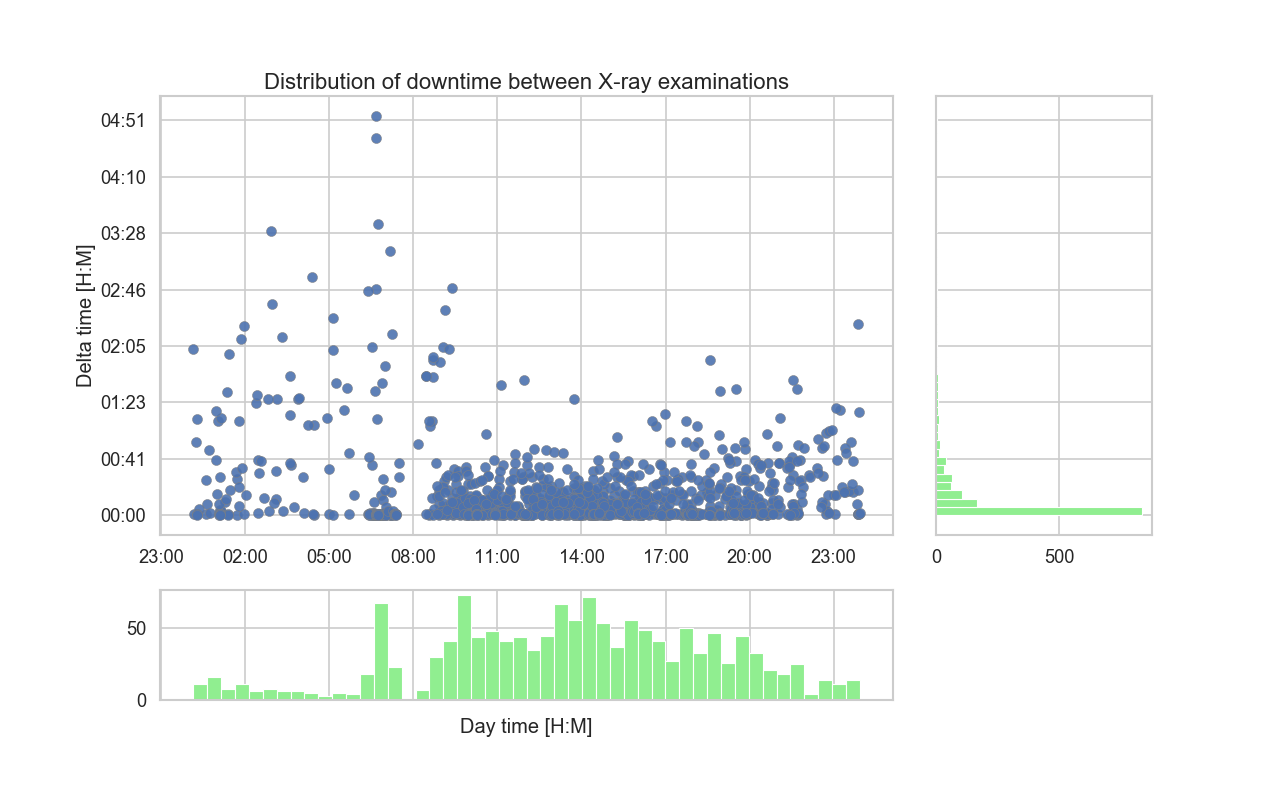}}
\end{figure}

Thereafter, each generated image is assigned zero or up to eight pathologies, based on the a-priority probabilities derived from the disease prevalence in the hospital. 
Finally, a model of a radiologist was set up, simultaneously working through the worklist by reporting CXRs and with a speed matching our CXR reporting time analysis. We sampled the reporting speed from our discrete distribution (Figure \ref{fig:DistributionReporting}), that included not only the raw reporting speed for a CXR but also other factors including pauses or interruptions due to phone calls.
\begin{figure}[htbp]
	\floatconts
	{fig:DistributionReporting}
	{\caption{Discrete distribution of chest X-ray reporting times by radiologists. The X-axis shows the day time in 24-hour format and the Y-axis shows the calculated time deltas between two CXR reports. The histogram in X- and Y- direction is shown in green.}}
	{\includegraphics[width=1\linewidth]{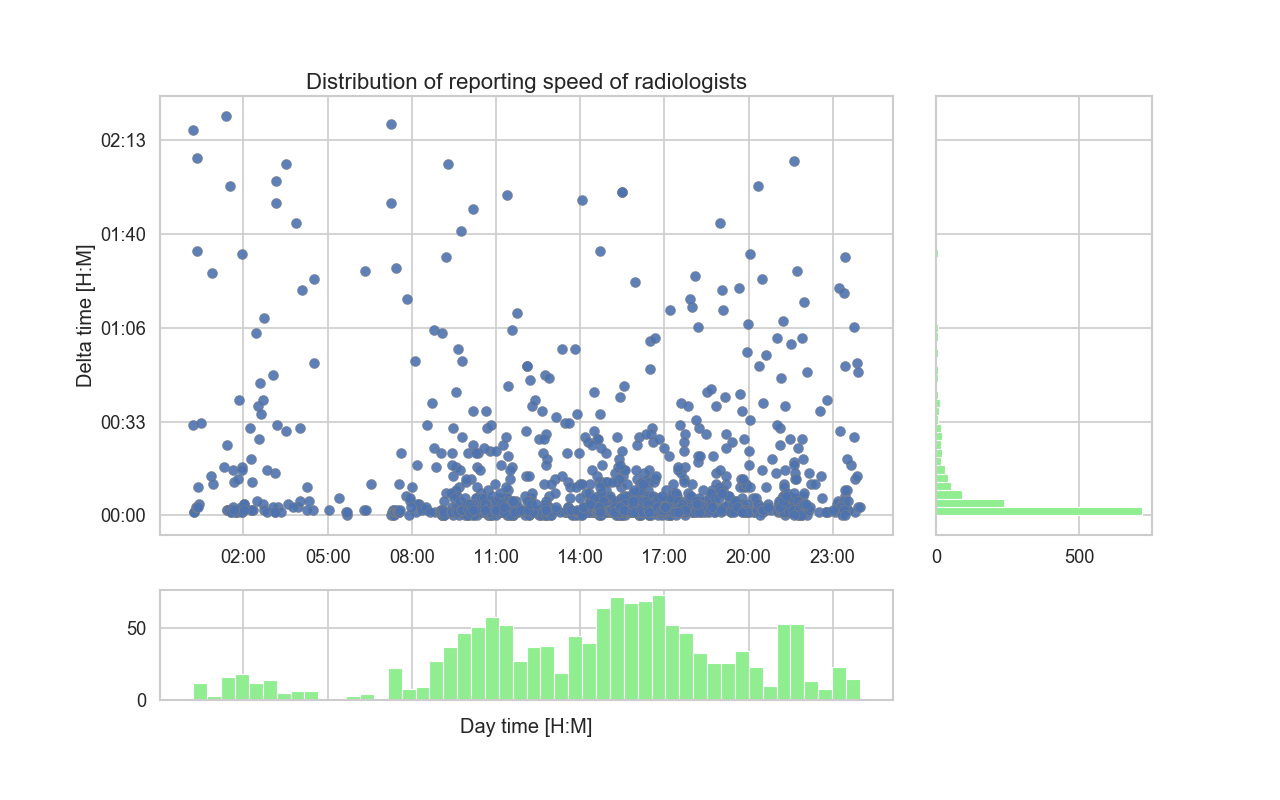}}
\end{figure}

This simulation models the current FIFO reporting and is therefore similar to the current clinical workflow. For the smart worklist prioritization, we included our AI model directly after the CXR generation. The AI model predicts for each CXR whether a finding is present or not before it is sorted into the worklist.

By automatically predicting the presence of all eight pathological findings an urgency level can be assigned according to Table \ref{tab:Finding}. Depending on the estimated level, the image is inserted into the existing worklist, taking prior images with a similar or higher level into account. The rearranged worklist is processed by the same radiologist model as in the FIFO simulation. 

To counteract the result of false negative predictions (i.e. sorting positive examinations to the end of the worklist), we propose a thresholding of the maximum waiting time. After an examination is longer than this maximum waiting time in the worklist, it is assigned with the highest priority and moved to the front of the worklist. While this should help to reduce the problem caused by false negative predictions (i.e. dangerously long maximum RTATs), it will also counteract our original goal of reducing the average RTAT for critical findings. 

All methods were tested using a Monte-Carlo simulation over 11.000 days with 24 hours of clinical routine, covering the generation of about 1.000.000 CXRs. Furthermore, the worklist was completely processed to zero once every 24 hours in all simulations. In our evaluation, we compared the average and the maximum RTAT of the simulations.

\section{Results}
\subsection{Pathology Distribution}
The analysis of pathology distribution at a university hospital was extracted by manually annotating 600 CXR reports from August 2016 and February 2019 by an expert radiologist. Due to a mainly stationary patient collective from a hospital of maximum care the portion of CXR without pathological findings was only 33\%. The prevalence of the most critical finding ``pneumothorax'' was around 5\%. Results are demonstrated in Table \ref{tab:Finding}.

\begin{table}[htbp]
	\floatconts
	{tab:Finding}%
	{\caption{Finding prevalence in chest X-rays at the university hospital (approximation by 600 samples from August 2016 and February 2019). The table is ordered by finding urgency as defined by our expert radiologists.}}%
	{\begin{tabular}{l r r}
			\bfseries Finding & \bfseries Total count & \bfseries Prevalence [\%]\\
			Pneumothorax 		& 23 & 3.8\\
			Congestion	 		& 124 & 20.7\\
			Pleural effusion 	& 236 & 39.3\\
			Infiltrate 			& 100 & 16.7\\
			Atelectasis 		& 124 & 20.7\\
			Cardiomegaly 		& 117 & 19.5\\
			Mass 				& 38 & 6.3\\
			Foreign Object 		& 298 & 49.7\\
			Normal 				& 186 & 31.0
	\end{tabular}}
\end{table}

\subsection{CXR Generation and Reporting Time Analysis}
We used a total of 1408 examinations to determine a discrete distribution of CXR generation and radiologist’s reporting speed. The examinations were from two different and non-consecutive weeks from Monday 00:00 AM until Sunday 00:00 PM. For CXR generation speed, we used the creation timestamp of two consecutive CXRs to calculate the delta between their creation. The delta represents the acquisition rate of CXRs at the institution. We employed the same method for the reporting speed. Here we used the report finalization timestamp to determine the delta between two CXRs. Afterwards, we removed all deltas greater than 2h 30min, to ensure outliers are only found in one of the two distributions.

\subsection{Hospital’s Report Turnaround Time Analysis}
The average RTAT for a CXR, measured over two different and non-consecutive weeks (1408 examinations), was 80 min with a range between 1 min and 1041 min. Assuming that a CXR report by an experienced radiologist will only take several minutes, this range in reporting time can be explained by different external influences, such as night shifts, change of shifts, working breaks or backlogs.

\subsection{Selecting CNN operation point}
Before running our workflow experiments, we evaluated in an initial experiment the optimal operation point for our CNN to reduce the average RTAT. 
When employing CNN multi-label classification, a threshold for every pathology must be defined in order to derive a binary classification (i.e. finding present or not) from the continuous response of the model. This corresponds to the selection of an operation point on the ROC curve. While an exhaustive valuation of different thresholds for all pathologies is computationally prohibitive, we focused on pneumothorax only (the most critical finding in our setting). Here, we estimated the average RTAT for different operating points by sampling the ROC curve at different false positive rates (FPR).
As shown in Figure \ref{fig:OptimalOP}, higher FPR reduces the effect of smart worklist prioritization to almost zero -- i.e. almost all examinations are rated urgent. While the other extreme (i.e. low FPR), can have no effect either, if almost all images are rated non-urgent.
Figure \ref{fig:OptimalOP} also shows that the optimal operation point to reduce the average RTAT is at a FPR of 0.05. For this operation point, we show in Table \ref{tab:operationPoint} the corresponding true positive, false negative and true negative rates. 

As shown in Table \ref{tab:operationPoint}, the optimal operation point to reduce the average RTAT still has a moderate false negative rate (FNR) of around 0.20 for most findings. The higher the FNR, the more likely it is that individual examinations with critical findings will be sorted to the end of the worklist. Hence, we selected a second operation point with a low FNR of 0.05 to evaluate if this can help to reduce the maximum RTAT. Table \ref{tab:operationPoint} shows the corresponding false positive, true negative and true positive rates.

\begin{figure}[htbp]
	\floatconts
	{fig:OptimalOP}
	{\caption{Optimal operation point simulation for the artificial intelligence algorithm. To find the optimal operation point for reducing the average report turnaround time (RTAT) for critical findings, we run multiple simulations with different false positive rates between zero and one.}}
	{\includegraphics[width=1.0\linewidth]{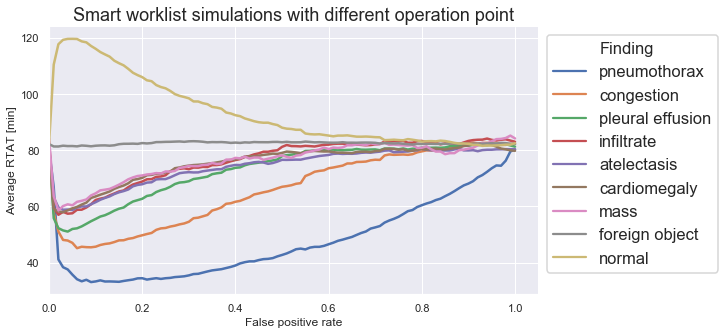}}
\end{figure}

\begin{table}[htbp]
	\floatconts
	{tab:operationPoint}%
	{\caption{Convolution neural network operation points. First column shows the operation point for best average report turnaround time (RTAT) reduction, while the second column shows the operation point for a low false negative rate (FNR) of 5\% (i.e. reducing the likelihood of dangerously long RTATs for critical findings).}}
	{ \begin{tabular}{l r r r | r r r}
			& \multicolumn{3}{c}{FPR = 0.05} & \multicolumn{3}{c}{FNR = 0.05}\\
			\bfseries Finding 	& \bfseries TPR & \bfseries FNR  & \bfseries TNR & \bfseries TPR & \bfseries FPR & \bfseries TNR \\
			Pneumothorax 		& 0.82 & 0.18 & 0.95 & 0.95 & 0.20 & 0.80 \\
			Congestion 			& 0.71 & 0.29 & 0.95 & 0.95 & 0.24 & 0.76 \\
			Pleural effusion 	& 0.86 & 0.14 & 0.95 & 0.95 & 0.21 & 0.79 \\
			Infiltrate 			& 0.75 & 0.25 & 0.95 & 0.95 & 0.27 & 0.73 \\
			Atelectasis 		& 0.61 & 0.39 & 0.95 & 0.95 & 0.39 & 0.61 \\
			Cardiomegaly 		& 0.75 & 0.25 & 0.95 & 0.95 & 0.18 & 0.82 \\
			Mass 				& 0.51 & 0.49 & 0.95 & 0.95 & 0.72 & 0.28 \\
			Foreign Object 		& 0.51 & 0.49 & 0.95 & 0.95 & 0.78 & 0.22
	\end{tabular}}
\end{table}

\subsection{Workflow Simulations}
Figure \ref{fig:Results} summarizes the effect of all four simulations (i.e. FIFO, Prio-lowFNR, Prio-lowFPR, Prio-MAXwaiting) on the RTAT. For the simulation Prio-lowFPR and Prio-MAXwaiting, we used the optimal operation point to reduce the average RTAT as shown in Table \ref{tab:operationPoint}. 
The average RTAT for critical findings was significantly reduced in the Prio-lowFPR simulation compared to the FIFO simulation –- e.g. pneumothorax: 35.6 min vs. 80.1 min, congestion: 45.3 min vs. 80.5 min, pleural effusion: 54.6 min vs. 80.5 min. As expected, an increase of average RTAT was only reported for normal examinations with a significant increase of the average RTAT from 80.2 min to 113.9 min. At the same time, however, the maximum RTAT in the Prio-lowFPR simulation increased compared to the FIFO simulation for all eight findings (e.g. pneumothorax: 1178 min vs. 890 min), as some examinations were predicted as false negative and sorted to the end of the worklist. The low FNR of 0.05 in Prio-lowFNR did not help to reduce the maximum RTAT (e.g. pneumothorax: 1293 min vs. 1178 min).

In the Prio-MAXwaiting simulation, we countered the false negative prediction problem by using a maximum waiting time and reduced the maximum RTAT for most findings (e.g. pneumothorax from 1178 min to 979 min). While, the average RTAT was only slightly higher than the Prio-lowFPR simulation (e.g. pneumothorax: 38.5 min vs. 35.6 min).

Finally, we also simulated the upper limit for a smart worklist prioritization by virtually employing a perfect classification algorithm (Perfect) with a true positive and true negative rate of 1. Table \ref{tab:Compare} shows the comparison with the other four simulations. For pneumothorax, the Prio-MAXwaiting average RTAT is only 8.3 min slower compared to the Perfect-simulation while FIFO is 49.8 min slower.

\paragraph*{Statistical Analysis}
The predictive performance of the CNN was assessed by using the area under the receiver operating characteristic curve (AUC). The shown AUC results are averaged over a 10-fold resampling.

We used the Welch’s t-test for the significant assessment of our smart worklist prioritization. First, we simulated a null distribution for the RTAT where examinations are sorted by the FIFO principle (i.e. random order). Secondly, we simulated an alternative distribution with worklist prioritization. Both distributions are then used to determine whether the average RTAT for each finding has changed significantly by calculating the p-value with the Welch t-test. Each distribution is simulated with a sample size of 1.000.000 examinations and the significant level is set to 0.05.
For all findings except ``foreign object'', we calculated a p $< 0.0001$. Hence, proving a significant change in the average RTAT.

\begin{figure}[htbp]
	\floatconts
	{fig:Results}
	{\caption{Report turnaround times (RTAT) for all eight pathological findings as well as for normal examinations on the basis of four different simulations: FIFO (green), Prio-lowFNR (yellow), Prio-lowFPR (purple) and Prio-MAXwaiting (red) with a maximum waiting time (light purple). The green triangles mark the average RTAT, while the vertical lines mark the median RTAT. On the right side, the maximum RTAT for each simulation and finding is shown.}}
	{\includegraphics[width=1.0\linewidth]{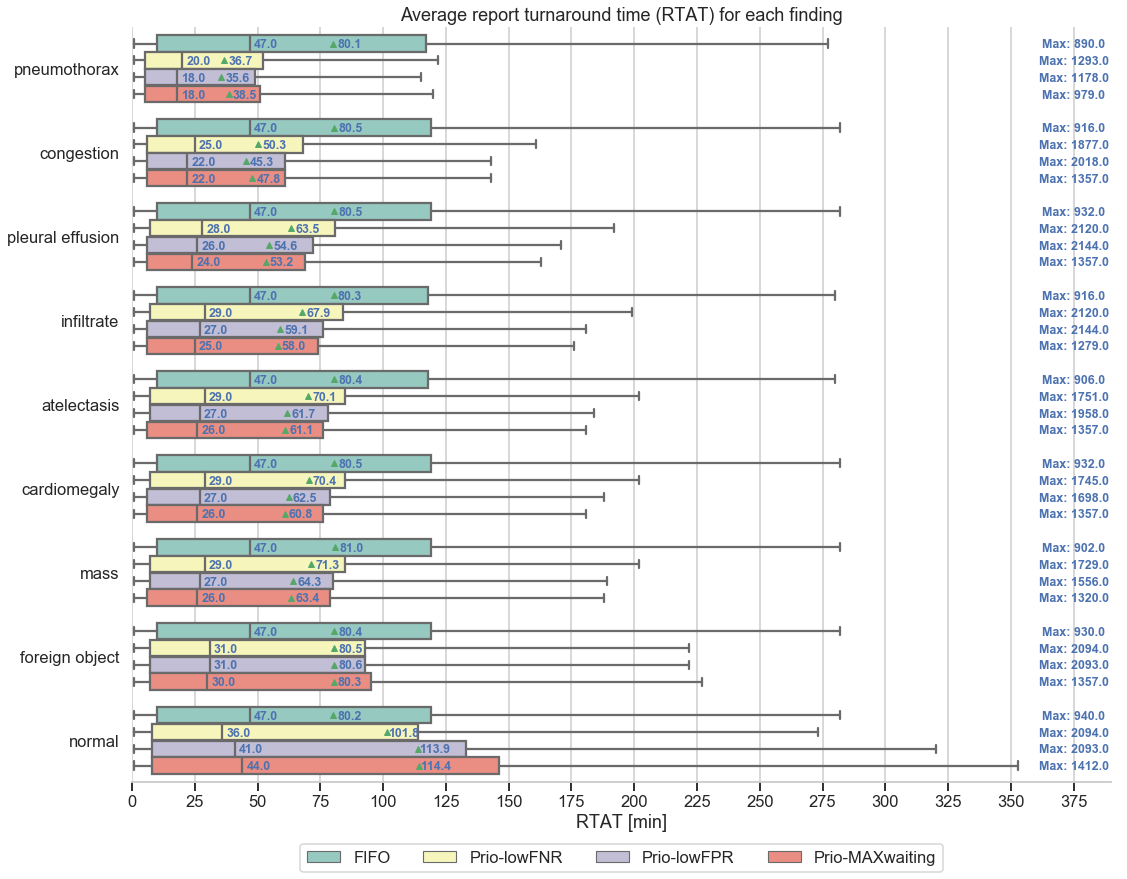}}
\end{figure}

\begin{table}[htbp]
	\floatconts
	{tab:Compare}%
	{\caption{Comparison of all four simulations (i.e. FIFO, Prio-lowFNR, Prio-lowFPR, Prio-MAXwaiting) with a perfect classification algorithm simulation (i.e. Perfect).}}%
	{{\footnotesize 
		\begin{tabular}{l r r r r r}
			\bfseries Finding 	& \bfseries FIFO & \bfseries Prio-lowFNR  & \bfseries Prio-lowFPR & \bfseries Prio-MAXwaiting & \bfseries Perfect\\
			Pneumothorax & 80.1 / 890 & 36.7 / 1293 & 35.6 / 1178 & 38.5 / \enspace979 &\enspace30.3 / \enspace320\\
			Congestion 			& 80.5 / 916 & 50.3 / 1877 & 45.3 / 2018 & 47.8 / 1357 &\enspace35.2 / \enspace510\\
			Pleural effusion 	& 80.5 / 932 & 63.5 / 2120 & 54.6 / 2144 & 53.2 / 1357 &\enspace45.4 / 1016\\
			Infiltrate 			& 80.3 / 916 & 67.9 / 2120 & 59.1 / 2144 & 58.0 / 1279 &\enspace49.8 / 1110\\
			Atelectasis 		& 80.4 / 906 & 70.1 / 1751 & 61.7 / 1958 & 61.1 / 1357 &\enspace51.4 / 1361\\
			Cardiomegaly 		& 80.5 / 932 & 70.4 / 1745 & 62.5 / 1698 & 60.8 / 1357 &\enspace52.2 / 1332\\
			Mass 				& 81.0 / 902 & 71.3 / 1729 & 64.3 / 1556 & 63.4 / 1320 &\enspace52.8 / 1301\\
			Foreign Object 		& 80.4 / 930 & 80.5 / 2094 & 80.6 / 2093 & 80.3 / 1357 & 80.7 / 2053\\
			Normal 				& 80.2 / 940 & 101.8 / 2094 & 113.9 / 2093 & 114.4 / 1412 & 131.5 / 2087
	\end{tabular}}}
\end{table}

\section{Discussion}
Our clinical workflow simulations demonstrated that a significant reduction of average RTAT for critical findings in CXRs can be achieved by a smart worklist prioritization using artificial intelligence. Furthermore, we showed that the problem of false negative predictions of an artificial intelligence system can be significantly reduced by introducing a maximum waiting time.

This was proven in a realistic clinical scenario, as all simulations were based on representative retrospective data from the university hospital. By extracting discrete distributions of CXR acquisition rate as well as radiologist’s reporting time, the temporal sequence of a working day could be recreated precisely.

As in other application areas, the question is what error rates we can ethically and legally tolerate before artificial intelligence can be used in patient care. For smart worklist prioritization, we have shown that we can easily reduce the average RTAT at the expense of individual cases that are classified as false negatives and therefore reported much later than the current FIFO principle. While it was questionable whether this overall improvement outweighed the risk of delayed reporting of individual cases, we have shown in our Prio-MAXwaiting simulation that the definition of a maximum waiting time, after which all examinations are assigned the highest priority, solves this problem. For the most critical finding (i.e. pneumothorax), the maximum RTAT was reduced to the current standard while preserving the significant reduction of the average RTAT.

The comparison in Table \ref{tab:Compare} shows that state-of-the-art convolutional neural networks can almost reach the upper limit of a smart worklist prioritization for the average RTAT. On the other hand, for the maximum RTAT, it reveals again the problem of false negative predictions. Ideally, a perfect classification algorithm could reduce the maximum RTAT to 320 min for pneumothorax, which is a substantial improvement over the standard maximum with 890 min.

Besides the possible improvement in diagnostic workflow by artificial intelligence, it should be stated that only a timely and reliable communication of the discovered findings from the radiologist to the referring clinician ensures that patients receive the clinical treatment they need.

Unlike previous publications [7] we included in- as well as outpatients, as in the daily reporting routine at the university hospital, all CXRs are sorted into one worklist. Furthermore, we observed substantially shorter backlogs of unreported examinations compared to published data from the United Kingdom.

In healthcare systems where patients and referring physicians are waiting for reports up to days or weeks, or have limited access to expert radiologists at all, the benefit of a smart worklist prioritization could obviously be greater than in countries with a well-developed health system. The longer the reporting backlogs, the more likely it is that referring physicians will try to rule out critical findings in CXRs themselves. This poses the risk that subtle findings with potentially large clinical impact (e.g. pneumothorax) are overlooked and that the discovery by a radiologist is postponed for a negligently long time period.

One limitation of our study is that the OpenI dataset, which our CNN was trained on, mainly included outpatients in contrast to the predominantly stationary patient collective of the hospital. Therefore, the performance of our algorithm, which was already strong compared to other publications [10], cannot be directly transferred to our hospital-specific patient collective and will most likely decrease. However, we note that the priority-based scheduling algorithm developed in this work is generic and can use any CNN that classifies chest X-ray pathologies. If the CNN classifier is improved, the scheduling algorithm will be directly benefited.

In the future, we want to include more pathologies and different degrees of manifestation to further improve the gain of a smart worklist prioritization. While we only focused on the eight most common findings in a CXR at the university hospital and ranked them according, a large atelectasis for example can put patients´ health more at risk than a small pleural effusion.

Overall, the application of smart worklist prioritization by artificial intelligence shows great potential to optimize clinical workflows and can significantly improve patient safety in the future. Our clinical workflow simulations suggest that triaging tools should be customized on the basis of local clinical circumstances and needs.


\bibliography{midl-samplebibliography}
1.) Reiner BI. Innovation Opportunities in Critical Results Communication: Theoretical Concepts. Journal of digital imaging. 2013 Aug 1;26(4):605-9.\\
2.) Hanna D, Griswold P, Leape LL, Bates DW. Communicating critical test results: safe practice recommendations. The Joint Commission Journal on Quality and Patient Safety. 2005 Feb 1;31(2):68-80.\\
3.) Singh H, Arora HS, Vij MS, Rao R, Khan MM, Petersen LA. Communication outcomes of critical imaging results in a computerized notification system. Journal of the American Medical Informatics Association. 2007 Jul 1;14(4):459-66.\\
4.) Berlin L. Statute of limitations and the continuum of care doctrine. American Journal of Roentgenology. 2001 Nov;177(5):1011-6.\\
5.) The Joint Commission. 2020 National Patient Safety Goals. Accessed \\November 5, 2019. Available from: \url{http://www.jointcommission.org/standards_information/npsgs.aspx}\\
6.) Rachh P, Levey AO, Lemmon A, Marinescu A, Auffermann WF, Haycook D, Berkowitz EA. Reducing STAT portable chest radiograph turnaround times: a pilot study. Current problems in diagnostic radiology. 2018 May 1;47(3):156-60.\\
7.) Gaskin CM, Patrie JT, Hanshew MD, Boatman DM, McWey RP. Impact of a reading priority scoring system on the prioritization of examination interpretations. American Journal of Roentgenology. 2016 May;206(5):1031-9.\\ 
8.) Yaniv G, Kuperberg A, Walach E. Deep learning algorithm for optimizing critical findings report turnaround time. In SIIM (Society for Imaging Informatics in Medicine) Annual Meeting 2018.\\
9.) Annarumma M, Withey SJ, Bakewell RJ, Pesce E, Goh V, Montana G. Automated triaging of adult chest radiographs with deep artificial neural networks. Radiology. 2019 Jan 22;291(1):196-202.\\
10.) Baltruschat IM, Nickisch H, Grass M, Knopp T, Saalbach A. Comparison of deep learning approaches for multi-label chest X-ray classification. Scientific reports. 2019 Apr 23;9(1):6381.\\
11.) Baltruschat IM, Steinmeister L, Ittrich H, Adam G, Nickisch H, Saalbach A, von Berg J, Grass M, Knopp T. When does Bone Suppression and Lung Field Segmentation Improve Chest X-Ray Disease Classification?. In2019 IEEE 16th International Symposium on Biomedical Imaging (ISBI 2019) 2019 Apr 8 (pp. 1362-1366). IEEE.\\
12.) Wang X, Peng Y, Lu L, Lu Z, Bagheri M, Summers RM. Chestx-ray8: Hospital-scale chest x-ray database and benchmarks on weakly-supervised classification and localization of common thorax diseases. InProceedings of the IEEE conference on computer vision and pattern recognition 2017 (pp. 2097-2106).\\
13.) Demner-Fushman D, Kohli MD, Rosenman MB, Shooshan SE, Rodriguez L, Antani S, Thoma GR, McDonald CJ. Preparing a collection of radiology examinations for distribution and retrieval. Journal of the American Medical Informatics Association. 2015 Jul 1;23(2):304-10.\\
14.) Gooßen A, Deshpande H, Harder T, Schwab E, Baltruschat I, Mabotuwana T, Cross N, Saalbach A. Deep Learning for Pneumothorax Detection and Localization in Chest Radiographs. arXiv preprint arXiv:1907.07324. 2019 Jul 16.\\
15.) Ondategui-Parra S, Bhagwat JG, Zou KH, Gogate A, Intriere LA, Kelly P, Seltzer SE, Ros PR. Practice management performance indicators in academic radiology departments. Radiology. 2004 Dec;233(3):716-22.\\

\end{document}